\documentclass[
]{ceurart}

\usepackage{listings}
\usepackage{caption}
\usepackage{subcaption}
\begin{document}

\copyrightyear{2023}
\copyrightclause{Copyright for this paper by its authors.
  Use permitted under Creative Commons License Attribution 4.0   International (CC BY 4.0).}

\conference{CLEF 2023: Conference and Labs of the Evaluation Forum, September 18–21, 2023, Thessaloniki, Greece}

\title{Fraunhofer SIT at CheckThat! 2023: Tackling Classification Uncertainty Using Model Souping on the Example of Check-Worthiness Classification}

\title[mode=sub]{Notebook for the CheckThat! Lab at CLEF 2023}


\author[1]{Raphael Antonius Frick}[%
email=raphael.frick@sit.fraunhofer.de,
]
\cormark[1]
\fnmark[1]
\address[1]{Fraunhofer Institute for Secure Information Technology SIT | ATHENE - National Research Center for Applied Cybersecurity,
  Rheinstrasse 75, Darmstadt, 64295, Germany,\\ url=https://www.sit.fraunhofer.de/}

\author[1]{Inna Vogel}[%
email=inna.vogel@sit.fraunhofer.de,
]
\fnmark[1]

\author[1]{Jeong-Eun Choi}[%
email=jeong-eun.choi@sit.fraunhofer.de,
]
\fnmark[1]

\cortext[1]{Corresponding author.}

\begin{abstract}
  This paper describes the second-placed approach developed by the Fraunhofer SIT team in the CLEF-2023 CheckThat! lab Task 1B for English. Given a text snippet from a political debate, the aim of this task is to determine whether it should be assessed for check-worthiness. Detecting check-worthy statements aims to facilitate manual fact-checking efforts by prioritizing the claims that fact-checkers should consider first. It can also be considered as primary step of a fact-checking system. Our best-performing method took advantage of an ensemble classification scheme centered on Model Souping. When applied to the English data set, our submitted model achieved an overall $F_{1}$ score of 0.878 and was ranked as the second-best model in the competition.
\end{abstract}

\begin{keywords}
  check-worthiness detection \sep
  model souping \sep
  fact-checking \sep
  BERT \sep
  NER
\end{keywords}

\maketitle

\begingroup
\let\clearpage\relax
\section{Introduction}
Spreading fake news, misinformation and disinformation online and on social media has become a pressing political and social issue. To combat such false or misleading information, several manual fact-checking initiatives have been launched, such as: FactCheck.org\footnote{\url{http://www.factcheck.org}}, PolitiFact\footnote{\url{http://www.politifact.com}} or Snopes\footnote{\url{https://www.snopes.com/fact-check/}}.  However, the speed at which misinformation is introduced and spread far outstrips the ability of any fact-checking organisation to verify information, so only the most salient claims are checked. Therefore, automatic identification of most worthy and prioritized claims for fact-checking can be very useful for human experts. The check-worthiness task can be considered as the first of three steps in the fact-checking pipeline, which traditionally consists of:

\begin{itemize}
	\item Detect check-worthy statements in a text. 	
	\item Retrieve claims that could be useful to fact-check and that have been verified in the past.
	\item Automated veracity estimation.
\end{itemize}

The primary step can be considered a filtering step in order to limit the overall computational effort required for the fact-checking pipeline and to reduce human resources. Further evidence to verify claims can be obtained from the internet by retrieving those that can be useful. To verify attention-worthy claims, trustworthy sources on the internet are searched. Finally, a decision can be made: whether the claim is factually true or not, based on the evidence gathered from the various sources \cite{checkthat_lab}.   

The CheckThat! Lab has been tackling this scientific problem for the past several years. This year, CheckThat! Lab offered two kinds of data for the check-worthiness subtask. For subtask 1A (multimodal), a text snippet (tweet) plus an image had to be assessed for check-worthiness. The aim of Task 1B (Multigenre) was to identify check-worthy statements from a tweet or a political debate/ speech transcription. Frauhofer SIT participated in Task 1A and 1B of the CLEF 2023 CheckThat! Lab Challenge for the English language. We achieved first place in Task 1A and second place in Task 1B. This paper describes the approach for Task 1B on identifying relevant claims in English political debates. 

Our approach takes advantage of ensemble learning to improve upon the uncertainty of single classifiers.
Since traditional stacking-based ensemble classifiers cause high computational overhead leading to long inference times, they are not always suitable for analyzing large data sets, especially data from social media.
Therefore, in this paper, we present an approach for detecting check-worthiness in texts that uses Model Souping to benefit from ensemble classification while consuming fewer resources and having low inference times.

The paper is structured as follows: Section 2 summarises the related work and some winning approaches from the last iterations of the challenge. In Section 3, we describe the data consisting of political debates that was provided by the CheckThat! Lab organizers. An overview of the tested and proposed approach is given in Section 4 along with their results on the respective data sets. The last section concludes our work with a brief discussion. 
\section{Related Work}

The initial check-worthiness detection methods were based on extracting meaningful features. Given U.S. presidential election transcripts, ClaimBuster \cite{Hassan2017} predicts check-worthiness by extracting a set of 6,615 features in total (sentiment, word count, tf-idf weighted bag-of-words, Part-of Speech tags, entity type), and used a SVM classifier for the prediction. Gencheva~et~al.~\cite{gencheva2017} extended the features used by ClaimBuster by including contextual features such as the sentence’s position, the size of a segment belonging to a speaker, topics or word embeddings. Using all features in combination with a neural network (FNN) outperformed the ClaimBuster version achieving a MAP of 0.427.

In the CheckThat! 2018 competition on check-worthiness detection Hansen et al. \cite{hansen2018} showed that a RNN with multiple word representations (word embeddings, POS tagging, and syntactic dependencies) could obtain state-of-the-art results for check-worthiness prediction. The authors later \cite{hansen2019} extended their work by applying weak supervision using a collection of unlabeled political speeches and showed significant improvements.
The objective of the CheckThat! challenge in 2021 was to determine which tweets within a set of COVID-19 related tweets are worth checking. The authors of the best performing model \cite{martinez2021} fine-tuned several pretrained transformer models. BERTweet achieved the best results (MAP 0.849 on the development set), a model that was trained on 850 million English tweets and 23 million COVID-19 related English tweets using RoBERTa. Savchev \cite{airational} experimented in the CheckThat! 2022 competition with three different pre-trained transformer models: BERT, DistilBERT and RoBERTa. Back translation (English tweets were translated to French and back to English) was applied to increase the training set. The best results (F1 0.90, Accuracy 0.85), and thus the first place in the competitions, were achieved by combining data augmentation and the RoBERTa model.  
\section{Data Set Description}

The ChackThat! Lab data sets for subtask 1B cover the languages Arabic, English and Spanish. While we only participated in the English language variant of the task, the described approach can also be adapted for other languages. 

For the English task the data set consisted of political debates collected from the US presidential general election debates. Examples from the data set are shown in Table~1. 

\begin{table}[h]
\caption{Instances of check-worthy (Yes) and non-check-worthy (No) sentences for Task 1B} 
	\centering
	\resizebox{\columnwidth}{!}{%
		\begin{tabular}{clc}
			\hline
			\multicolumn{1}{l}{} & Instance                                                & \multicolumn{1}{l}{Class} \\ \hline
			1. & "And that means 98 percent of American families, 97 percent of small businesses, they will not see a tax increase." & Yes \\
			2. & I said we'd get tougher with child support and child support enforcement's up 50 percent.                           & Yes \\
			3.                   & But I'm not going to do that.                           & No                        \\
			4.                   & But the important thing is what are we going to do now? & No                        \\ \hline
		\end{tabular}%
	}
\label{tab:sentExamples}
\end{table}

The aim of Task 1B was to predict whether a text snippet from a political debate has to be assessed manually by an expert by estimating its check-worthiness. The data set was annotated by human labelers. The label distributions and data set split were provided by the organizers and are shown in Table~2. The "train" corpus consists of 16,876 entries. Each entry is labeled either "Yes" or "No" on whether it is worth fact-checking (YES) or not (No). The organizers have also provided a development set "dev" (5,625 entries), a development test set "dev test" (1,032 entries), and a test set with 318 statements. As it can be seen, the data set is highly imbalanced with about a quarter of the sentences being check-worthy. This is also due to the fact that attention-worthy sentences occur less frequently in the text than non-check-worthy sentences.

\begin{table}[h]
	\centering
	\caption{Class distribution of the CheckThat! Lab 2023 task 1B English data set}
	\label{tab:class_destribution}
	\begin{tabular}{llll}
		\hline
		& Total  & Yes   & No     \\ \hline
		Train    & 16,876 & 4,058 & 12,818 \\
		Dev      & 5,625  & 1,355 & 4,270  \\
		Dev Test & 1,032  & 238   & 794    \\
		Test     & 318    & 108   & 210    \\ \hline
		Sum      & 23,851 & 5,759 & 18,092 \\ \hline
	\end{tabular}
\end{table}

\section{Methodology and Results}

Text data from social media and messenger applications such as Twitter and Telegram, news and blogging websites, and transcribed political debates may contain incorrect information that needs to be subjected to manual review by an expert.
Hereby, texts of interest are those that contain asserted facts that can be proven or disproven.
To identify if a text is check-worthy, three approaches have been tested as part of the CheckThat! 2023 competetion: an \emph{estimation using named entities}, a method \emph{combining the named entity recognition with BERT} and the final solution consisting of an \emph{ensemble classifier based on Model Souping}.
In the following, the three approaches will be described and their performance discussed.

\subsection{Estimation Using Named Entity Analysis}
Facts can often be expressed using named entities, such as names (person / coorperation / location / event / objects) or numbers (cardinals / ordinals / quantities) and dates.
A thorough examination of the train partitioning of the data set revealed that samples classified as check-worthy had a higher use of named entities than those classified as not worthy of reviewing.
Using Flair \cite{akbik2019flair}, a named entity recognition model pre-trained on the OntoNotes data set\cite{Onto}, the named entities within each of the provided text snippets were extracted and categorized.
Hereby, check-worthy texts contained on average $1.679$ named entities, whereby non-attention-worthy texts featured only $0.662$ named entities on average.
Further analysis showed that in addition to the number of named entities featured, the type of entities also varied between the two classes.
Figure \ref{fig:type_dist} showcases the distribution of named entity types pre-grouped by similarity.
The parent type \emph{NUM} consists of ordinal numbers, cardinal numbers, quantities, percentages, and money, while \emph{DATE} consists of time and dates. \emph{GPE} consists of nationalities, countries, and states, and \emph{LOC} consists of places and events.
\emph{PER} and \emph{ORG} remained self-contained. 
The distribution shows that texts worth examining often contain numbers and counts, while nationalities, countries and states are found less frequently.

\begin{figure}
        \caption{Normalized distribution of named entity types}
    
     \centering
     \begin{subfigure}[b]{0.48\textwidth}
         \centering
         \includegraphics[width=\textwidth]{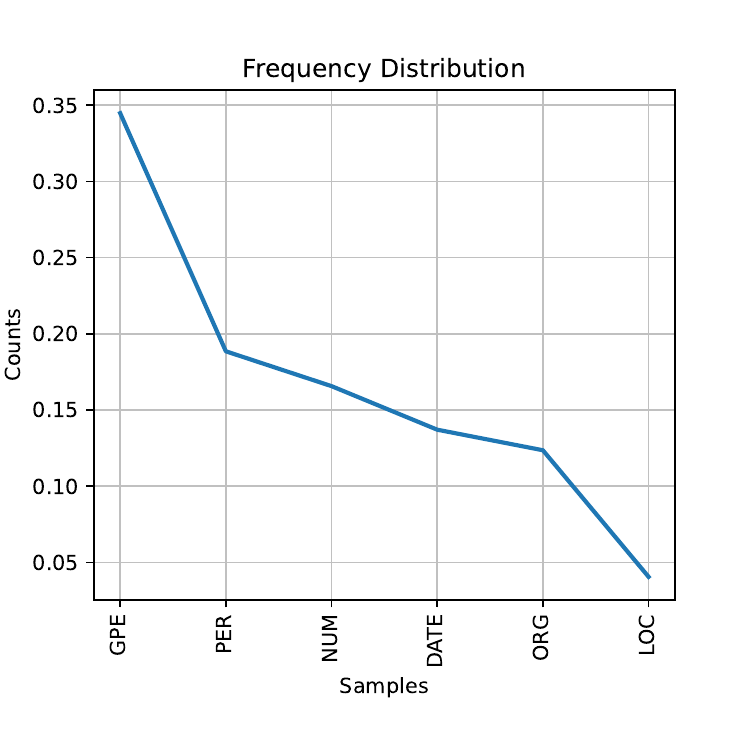}
         \caption{In  \emph{non-check-worthy} texts}
         \end{subfigure}
     \hfill
     \begin{subfigure}[b]{0.48\textwidth}
         \centering
         \includegraphics[width=\textwidth]{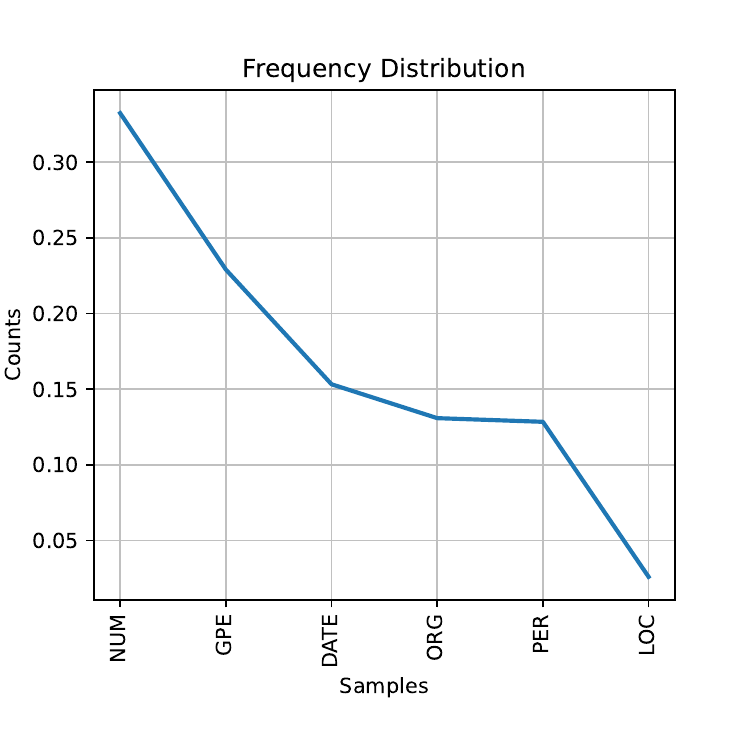}
         \caption{In  \emph{check-worthy} texts}
         \end{subfigure}
         \label{fig:type_dist}
\end{figure}

The resulting information was then used to train a classifier, namely a logistic regression model, using the number of a given parent type as input.
As indicated in Table \ref{tab:results}, the model was able to achieve medium to high accuracies, especially when classifying dev and dev-test split of the data set.
In comparison, however, the $F_1$-values are very low, making the model unsuitable for real-world applications.
One reason for this lies in the imbalanced class distribution within the data set.
Another one is that analyzing the occurrence of named entities alone does not provide enough information for a precise estimation.
A text mentioning a large number of named entities but which is written in a subjective tone and expresses a personal opinion is not worthy of review. 
Thus, to mitigate this problem, contextual information must be analyzed as well.

\begin{table}[]
\caption{Scores achieved by each model on each data set. $A$ refers to the accuracy score, $P$ to the precision score, $R$ to the recall score, and $F_1$ to the $F_1$-score. The character $d$ denotes the dev data split, $dt$ the dev-test data split, and $t$ the test split of the data set.}

\resizebox{\columnwidth}{!}{
\begin{tabular}{|c|c|c|c|c|c|c|}
\hline
\multirow{2}{*}{} & \multirow{2}{*}{\begin{tabular}[c]{@{}l@{}}Logistic Regression \\ + NER\end{tabular}} & \multirow{2}{*}{\begin{tabular}[c]{@{}l@{}}BERT \\ + NER\end{tabular}} & \multirow{2}{*}{BERT A} & \multirow{2}{*}{BERT B} & \multirow{2}{*}{BERT C} & \multirow{2}{*}{\begin{tabular}[c]{@{}l@{}}Model Souping \\ + BERT\end{tabular}} \\
                  &                                                                                       &                                                                        &                         &                         &                         &                                                                                  \\ \hline

$A_d$     & 0.7909                    &  0.8796          &   0.8728    &   0.8764     &    0.8565    &      0.8670                \\ \hline
$P_d$     & 0.6751                    &   0.7834         &   0.7524    &   0.7248     &    0.6608    &      0.6849                \\ \hline
$R_d$     & 0.2546                    &    0.6915        &   0.7041    &   0.7852     &    0.8310    &        0.8295              \\ \hline
$F1_d$    & 0.3697                    &   0.7346         &   0.7274    &   0.7538     &    0.7362    &        0.7503              \\ \hline
$A_{dt}$  & 0.8430                    &    0.9554        &   0.9690    &   0.9729     &    0.9680    &       0.9709               \\ \hline
$P_{dt}$  & 0.8333                    &       0.9444     &   0.9558    &   0.9303     &    0.8958     &      0.9094                \\ \hline
$R_{dt}$  & 0.3992                    &     0.8571       &   0.9076    &   0.9538      &   0.9748      &      0.9706                \\ \hline
$F1_{dt}$ & 0.5398                    &    0.8987        &   0.9310    &   0.9419      &   0.9336      &      0.9390                \\ \hline
$A_t$     & 0.6981                    &  0.8711          &   0.8710     &   0.9308     &   0.9308    &               0.9214       \\ \hline
$P_t$     & 0.7727                    &     0.9855       &    0.9351    &    0.9674    &  0.9216      &         0.9278             \\ \hline
$R_t$     & 0.1574                    &    0.6296        &  0.6667      &   0.8241     &   0.8704     &             0.8333         \\ \hline
$F1_t$    & 0.2615                    &   0.7684         &    0.7784    &  0.8900      &  0.8952      &          \textbf{0.8780}            \\ \hline
\end{tabular}
}
\label{tab:results}
\end{table}

\subsection{Combining the Analysis of Named Entities with Language Models}

To include additional information about the context, the second attempt combined the named entity recognition with a language model.
Here, BERT\cite{devlin2019bert} was fine-tuned on data, in which the named entities found in the previous step were exchanged with special tokens reflecting their respective named entity type (see Table \ref{tab:NERsentExamples}).
For this, the tokenizer was modified to contain the six additional tokens \emph{<NUM>, <DATE>, <LOC>, <GPE>, <PER>}, and \emph{<ORG>}.

\begin{table}[h]
\caption{Examples of a named entity extraction in check-worthy (Yes), and non-check-worthy (No) sentences} 
	\centering
	\resizebox{\columnwidth}{!}{%
		\begin{tabular}{clc}
			\hline
			\multicolumn{1}{l}{} & Instance                                                & \multicolumn{1}{l}{Class} \\ \hline
			1. & "And that means <98 percent, NUM> of <American, GPE> families, <97 percent, NUM> of small businesses, they will not see a tax increase." & Yes \\
			2. & I said we'd get tougher with child support and child support enforcement's up <50 percent, NUM>.                           & Yes \\
			3.                   & But I'm not going to do that.                           & No                        \\
			4.                   & But the important thing is what are we going to do now? & No                        \\ \hline
		\end{tabular}%
	}
\label{tab:NERsentExamples}
\end{table}

During training, an optimizer based on Adam\cite{kingma2017adam} was utilized to leverage from the adaptive learning rate mechanism. 
A learning rate of $0.0004$ was chosen as the initial learning rate. The model was fine-tuned in $5$ epochs with a batch size of $24$.
Model checkpoints were used to keep only the model checkpoint, that performed best on the dev split of the competition data set.

Table \ref{tab:results} shows, that by combining a named entity recognition with a language model such as BERT, the performance can be further increased.

\subsection{Final Solution Using Ensemble Learning Based on Model Souping}

To compare the hybrid method with a fully data-driven approach, fine-tuning was performed using solely the raw text data.
Again, a BERT model was chosen and fine-tuned using the configuration described in the previous section.

Training the model several times with different seeds showed large differences in performance (see BERT A, BERT B, and BERT C in Table \ref{tab:results}). 
The reason for this is that the initial weights of the model are initialized differently depending on the set seed.
The same applies to the way the training split is shuffled after each epoch.
As a result, individual models converge differently and are able to find different local minima, resulting in a sometimes good or less good performance.
Unable to determine which seed maximizes performance on the validation, and test sets, it is common to take advantage of ensemble learning.

There exist several different approaches to perform ensemble classification, such as \emph{bagging}, \emph{boosting}, and \emph{stacking}\cite{ensemble}.
In each of the methods, individually trained classifiers called weak classifiers are combined to improve the classification uncertainty.
The main disadvantage of ensemble learners, however, lies within their computational efficiency during inference.
In particular, stacking-based ensemble classifiers, which consist of a combination of $N$ models providing an initial prediction and a meta classifier taking these to form a final decision output, require the inference of $N+1$ models.
As such, ensemble classification may not be applicable in real-world applications, in which large amounts of data need to be assessed in a timely matter using as less computational resources as possible. 

To compensate for these problems, \emph{Model Souping} as proposed by Wortsman et al. can be applied \cite{wortsman2022model}.
Model Souping removes the requirement of having multiple weak classifiers and a meta-classifier by providing a single master-model that is used during inference.
Master-models can be built by taking the trained weights of each individual classifier and combining them by averaging, weighted averaging, or using a feedback loop.
Initial tests with image and text classification tasks showed improved performance while maintaining resource efficiency.
It should be noted, however, that Model Souping can only be applied with models sharing the same architecture.

In this paper, we took advantage of Model Soups that adaptively adjust the influence of each individual model in the master model based on the performance on the dev split of the data set.
Here, the fully data-driven models were used in favor of the hybrid models BERT with a named entity recognition due to their performance on the dev, and dev-test split.
By evaluating each of the three trained models on the dev set, their test loss values were retrieved. 
Based on them, their influence-score $I$ was calculated using the following formula:
\begin{equation}
    I = L_{test} / L_{total}
\end{equation}
Low-performing models should have a lower impact within the master model, whereas better-performing ones, should have a higher influence on the weights of the master model.
The influence value $I$ was then used to weight the trained weights of each model.  

While the ensemble classifier was not able to outperform the best individual classifier (BERT~C;~$F_1~=~0.8952$) on the test data set, it helped with balancing out results from models (BERT~A;~$F_1~=~0.7784$) suffering from low performance. 
It should be noted, however, that if all weak classifiers perform equally well on a particular data set, the performance gain will be negligible.

The approach based on Model Souping was used to classify the private test set of this year's CheckThat! competition. 
It was able to place second best.
Although it performed best among the three methods described, its capabilities in terms of explainability and transparency are limited due to the fact that it is a fully data-driven approach.
\section{Conclusion}

The detection of check-worthy texts can be seen as a first step to identifying false information spread on the internet. When used as a pre-filter, it can dramatically reduce the amount of data that needs to be manually reviewed by human experts. The paper provides a new way of detecting attention-worthy content using an ensemble classification scheme based on Model Souping. Experiments on the validation split and the private test set revealed that the proposed approach can be used to tackle the issue of classification uncertainty while reducing the computational overhead often associated with ensemble learning. The model was able to place second best in the competition with a $F_1$-score of $0.878$. Future work may consider applying weight adjustments using a feedback loop to better compensate for the misclassification of edge cases as well as introducing other means to achieve explainability and transparency.   
\section*{Acknowledgements}
This work was supported by the German Federal Ministry of Education and Research (BMBF) and the  Hessian Ministry of Higher Education, Research, Science and the Arts within their joint support of “ATHENE – CRISIS” and "Lernlabor Cybersicherheit" (LLCS).
\endgroup

\bibliography{clef23_template}

\begin{thebibliography}{13}
\expandafter\ifx\csname natexlab\endcsname\relax\def\natexlab#1{#1}\fi
\providecommand{\url}[1]{\texttt{#1}}
\providecommand{\href}[2]{#2}
\providecommand{\path}[1]{#1}
\providecommand{\DOIprefix}{doi:}
\providecommand{\ArXivprefix}{arXiv:}
\providecommand{\URLprefix}{URL: }
\providecommand{\Pubmedprefix}{pmid:}
\providecommand{\doi}[1]{\href{http://dx.doi.org/#1}{\path{#1}}}
\providecommand{\Pubmed}[1]{\href{pmid:#1}{\path{#1}}}
\providecommand{\bibinfo}[2]{#2}
\ifx\xfnm\relax \def\xfnm[#1]{\unskip,\space#1}\fi
\bibitem[{Barr{\'o}n-Cede{\~{n}}o et~al.(2023)Barr{\'o}n-Cede{\~{n}}o, Alam,
  Caselli, Da~San~Martino, Elsayed, Galassi, Haouari, Ruggeri, Struss, Nandi,
  Cheema, Azizov, and Nakov}]{checkthat_lab}
\bibinfo{author}{A.~Barr{\'o}n-Cede{\~{n}}o}, \bibinfo{author}{F.~Alam},
  \bibinfo{author}{T.~Caselli}, \bibinfo{author}{G.~Da~San~Martino},
  \bibinfo{author}{T.~Elsayed}, \bibinfo{author}{A.~Galassi},
  \bibinfo{author}{F.~Haouari}, \bibinfo{author}{F.~Ruggeri},
  \bibinfo{author}{J.~M. Struss}, \bibinfo{author}{R.~N. Nandi},
  \bibinfo{author}{G.~S. Cheema}, \bibinfo{author}{D.~Azizov},
  \bibinfo{author}{P.~Nakov},
\newblock \bibinfo{title}{The clef-2023 checkthat! lab: Checkworthiness,
  subjectivity, political bias, factuality, and authority},
\newblock in: \bibinfo{editor}{J.~Kamps}, \bibinfo{editor}{L.~Goeuriot},
  \bibinfo{editor}{F.~Crestani}, \bibinfo{editor}{M.~Maistro},
  \bibinfo{editor}{H.~Joho}, \bibinfo{editor}{B.~Davis},
  \bibinfo{editor}{C.~Gurrin}, \bibinfo{editor}{U.~Kruschwitz},
  \bibinfo{editor}{A.~Caputo} (Eds.), \bibinfo{booktitle}{Advances in
  Information Retrieval}, \bibinfo{publisher}{Springer Nature Switzerland},
  \bibinfo{address}{Cham}, \bibinfo{year}{2023}, pp. \bibinfo{pages}{506--517}.
\bibitem[{Hassan et~al.(2017)Hassan, Arslan, Li, and Tremayne}]{Hassan2017}
\bibinfo{author}{N.~Hassan}, \bibinfo{author}{F.~Arslan},
  \bibinfo{author}{C.~Li}, \bibinfo{author}{M.~Tremayne},
\newblock \bibinfo{title}{Toward automated fact-checking: Detecting
  check-worthy factual claims by claimbuster},
\newblock \bibinfo{journal}{Proceedings of the 23rd ACM SIGKDD International
  Conference on Knowledge Discovery and Data Mining}  (\bibinfo{year}{2017}).
\bibitem[{Gencheva et~al.(2017)Gencheva, Nakov, M{\`a}rquez,
  Barr{\'o}n-Cede{\~n}o, and Koychev}]{gencheva2017}
\bibinfo{author}{P.~Gencheva}, \bibinfo{author}{P.~Nakov},
  \bibinfo{author}{L.~M{\`a}rquez}, \bibinfo{author}{A.~Barr{\'o}n-Cede{\~n}o},
  \bibinfo{author}{I.~Koychev},
\newblock \bibinfo{title}{A context-aware approach for detecting worth-checking
  claims in political debates},
\newblock in: \bibinfo{booktitle}{Proceedings of the International Conference
  Recent Advances in Natural Language Processing, {RANLP} 2017},
  \bibinfo{publisher}{INCOMA Ltd.}, \bibinfo{address}{Varna, Bulgaria},
  \bibinfo{year}{2017}, pp. \bibinfo{pages}{267--276}. \URLprefix
  \url{https://doi.org/10.26615/978-954-452-049-6_037}.
  \DOIprefix\doi{10.26615/978-954-452-049-6_037}.
\bibitem[{Hansen et~al.(2018)Hansen, Hansen, Simonsen, and Lioma}]{hansen2018}
\bibinfo{author}{C.~Hansen}, \bibinfo{author}{C.~Hansen},
  \bibinfo{author}{J.~Simonsen}, \bibinfo{author}{C.~Lioma},
\newblock \bibinfo{title}{The copenhagen team participation in the
  check-worthiness task of the competition of automatic identification and
  verification of claims in political debates of the clef-2018 checkthat! lab},
\newblock in: \bibinfo{editor}{L.~{Cappellato }}, \bibinfo{editor}{N.~{Ferro
  }}, \bibinfo{editor}{J.~Nie}, \bibinfo{editor}{L.~Soulier} (Eds.),
  \bibinfo{booktitle}{CLEF 2018 Working Notes}, CEUR Workshop Proceedings,
  \bibinfo{publisher}{CEUR-WS.org}, \bibinfo{year}{2018}. \bibinfo{note}{19th
  Working Notes of CLEF Conference and Labs of the Evaluation Forum, CLEF 2018
  ; Conference date: 10-09-2018 Through 14-09-2018}.
\bibitem[{Hansen et~al.(2019)Hansen, Hansen, Simonsen, and Lioma}]{hansen2019}
\bibinfo{author}{C.~Hansen}, \bibinfo{author}{C.~Hansen},
  \bibinfo{author}{J.~G. Simonsen}, \bibinfo{author}{C.~Lioma},
\newblock \bibinfo{title}{Neural weakly supervised fact check-worthiness
  detection with contrastive sampling-based ranking loss},
\newblock in: \bibinfo{editor}{L.~Cappellato}, \bibinfo{editor}{N.~Ferro},
  \bibinfo{editor}{D.~E. Losada}, \bibinfo{editor}{H.~M{\"{u}}ller} (Eds.),
  \bibinfo{booktitle}{Working Notes of {CLEF} 2019 - Conference and Labs of the
  Evaluation Forum, Lugano, Switzerland, September 9-12, 2019}, volume
  \bibinfo{volume}{2380} of \textit{\bibinfo{series}{{CEUR} Workshop
  Proceedings}}, \bibinfo{publisher}{CEUR-WS.org}, \bibinfo{year}{2019}.
  \URLprefix \url{https://ceur-ws.org/Vol-2380/paper\_56.pdf}.
\bibitem[{Martinez{-}Rico et~al.(2021)Martinez{-}Rico, Mart{\'{\i}}nez{-}Romo,
  and Araujo}]{martinez2021}
\bibinfo{author}{J.~R. Martinez{-}Rico},
  \bibinfo{author}{J.~Mart{\'{\i}}nez{-}Romo}, \bibinfo{author}{L.~Araujo},
\newblock \bibinfo{title}{Nlp{\&}ir@uned at checkthat!~2021: Check-worthiness
  estimation and fake news detection using transformer models},
\newblock in: \bibinfo{editor}{G.~Faggioli}, \bibinfo{editor}{N.~Ferro},
  \bibinfo{editor}{A.~Joly}, \bibinfo{editor}{M.~Maistro},
  \bibinfo{editor}{F.~Piroi} (Eds.), \bibinfo{booktitle}{Proceedings of the
  Working Notes of {CLEF} 2021 - Conference and Labs of the Evaluation Forum,
  Bucharest, Romania, September 21st - to - 24th, 2021}, volume
  \bibinfo{volume}{2936} of \textit{\bibinfo{series}{{CEUR} Workshop
  Proceedings}}, \bibinfo{publisher}{CEUR-WS.org}, \bibinfo{year}{2021}, pp.
  \bibinfo{pages}{545--557}. \URLprefix
  \url{https://ceur-ws.org/Vol-2936/paper-44.pdf}.
\bibitem[{Savchev(2022)}]{airational}
\bibinfo{author}{A.~Savchev},
\newblock \bibinfo{title}{{AI} rational at checkthat!-2022: Using transformer
  models for tweet classification},
\newblock in: \bibinfo{editor}{G.~Faggioli}, \bibinfo{editor}{N.~Ferro},
  \bibinfo{editor}{A.~Hanbury}, \bibinfo{editor}{M.~Potthast} (Eds.),
  \bibinfo{booktitle}{Proceedings of the Working Notes of {CLEF} 2022 -
  Conference and Labs of the Evaluation Forum, Bologna, Italy, September 5th -
  to - 8th, 2022}, volume \bibinfo{volume}{3180} of
  \textit{\bibinfo{series}{{CEUR} Workshop Proceedings}},
  \bibinfo{publisher}{CEUR-WS.org}, \bibinfo{year}{2022}, pp.
  \bibinfo{pages}{656--659}. \URLprefix
  \url{https://ceur-ws.org/Vol-3180/paper-52.pdf}.
\bibitem[{Akbik et~al.(2019)Akbik, Bergmann, Blythe, Rasul, Schweter, and
  Vollgraf}]{akbik2019flair}
\bibinfo{author}{A.~Akbik}, \bibinfo{author}{T.~Bergmann},
  \bibinfo{author}{D.~Blythe}, \bibinfo{author}{K.~Rasul},
  \bibinfo{author}{S.~Schweter}, \bibinfo{author}{R.~Vollgraf},
\newblock \bibinfo{title}{{FLAIR}: An easy-to-use framework for
  state-of-the-art {NLP}},
\newblock in: \bibinfo{booktitle}{{NAACL} 2019, 2019 Annual Conference of the
  North American Chapter of the Association for Computational Linguistics
  (Demonstrations)}, \bibinfo{year}{2019}, pp. \bibinfo{pages}{54--59}.
\bibitem[{{Weischedel, Ralph} et~al.(2013){Weischedel, Ralph}, {Palmer,
  Martha}, {Marcus, Mitchell}, {Hovy, Eduard}, {Pradhan, Sameer}, {Ramshaw,
  Lance}, {Xue, Nianwen}, {Taylor, Ann}, {Kaufman, Jeff}, {Franchini,
  Michelle}, {El-Bachouti, Mohammed}, {Belvin, Robert}, and {Houston,
  Ann}}]{Onto}
\bibinfo{author}{{Weischedel, Ralph}}, \bibinfo{author}{{Palmer, Martha}},
  \bibinfo{author}{{Marcus, Mitchell}}, \bibinfo{author}{{Hovy, Eduard}},
  \bibinfo{author}{{Pradhan, Sameer}}, \bibinfo{author}{{Ramshaw, Lance}},
  \bibinfo{author}{{Xue, Nianwen}}, \bibinfo{author}{{Taylor, Ann}},
  \bibinfo{author}{{Kaufman, Jeff}}, \bibinfo{author}{{Franchini, Michelle}},
  \bibinfo{author}{{El-Bachouti, Mohammed}}, \bibinfo{author}{{Belvin,
  Robert}}, \bibinfo{author}{{Houston, Ann}}, \bibinfo{title}{Ontonotes release
  5.0}, \bibinfo{year}{2013}. \URLprefix
  \url{https://catalog.ldc.upenn.edu/LDC2013T19}.
  \DOIprefix\doi{10.35111/XMHB-2B84}.
\bibitem[{Devlin et~al.(2019)Devlin, Chang, Lee, and
  Toutanova}]{devlin2019bert}
\bibinfo{author}{J.~Devlin}, \bibinfo{author}{M.-W. Chang},
  \bibinfo{author}{K.~Lee}, \bibinfo{author}{K.~Toutanova},
  \bibinfo{title}{Bert: Pre-training of deep bidirectional transformers for
  language understanding}, \bibinfo{year}{2019}.
  \href{http://arxiv.org/abs/1810.04805}{{\tt arXiv:1810.04805}}.
\bibitem[{Kingma and Ba(2017)}]{kingma2017adam}
\bibinfo{author}{D.~P. Kingma}, \bibinfo{author}{J.~Ba}, \bibinfo{title}{Adam:
  A method for stochastic optimization}, \bibinfo{year}{2017}.
  \href{http://arxiv.org/abs/1412.6980}{{\tt arXiv:1412.6980}}.
\bibitem[{Bauer and Kohavi(1996)}]{ensemble}
\bibinfo{author}{E.~Bauer}, \bibinfo{author}{R.~Kohavi},
\newblock \bibinfo{title}{An empirical comparison of voting classification
  algorithms : Bagging, boosting, and variants},
\newblock \bibinfo{journal}{Machine Learning} \bibinfo{volume}{36}
  (\bibinfo{year}{1996}) \bibinfo{pages}{1--38}.
\bibitem[{Wortsman et~al.(2022)Wortsman, Ilharco, Gadre, Roelofs,
  Gontijo-Lopes, Morcos, Namkoong, Farhadi, Carmon, Kornblith, and
  Schmidt}]{wortsman2022model}
\bibinfo{author}{M.~Wortsman}, \bibinfo{author}{G.~Ilharco},
  \bibinfo{author}{S.~Y. Gadre}, \bibinfo{author}{R.~Roelofs},
  \bibinfo{author}{R.~Gontijo-Lopes}, \bibinfo{author}{A.~S. Morcos},
  \bibinfo{author}{H.~Namkoong}, \bibinfo{author}{A.~Farhadi},
  \bibinfo{author}{Y.~Carmon}, \bibinfo{author}{S.~Kornblith},
  \bibinfo{author}{L.~Schmidt}, \bibinfo{title}{Model soups: averaging weights
  of multiple fine-tuned models improves accuracy without increasing inference
  time}, \bibinfo{year}{2022}. \href{http://arxiv.org/abs/2203.05482}{{\tt
  arXiv:2203.05482}}.

\end{thebibliography}

\appendix

\end{document}